\documentclass[runningheads]{llncs}
\usepackage{graphicx}
\usepackage{multirow}
\usepackage{subfigure}
\usepackage{multirow}
\usepackage{colortbl,xcolor}
\definecolor{Gray}{gray}{0.8}
\usepackage{amsmath,amssymb} % define this before the line numbering.
\usepackage{color}
\usepackage{color}
\def\W{\mathbf{W}}
\def\A{\mathbf{A}}

\def\ie#1{\textit{i.e.}{#1}}
\def\eg#1{\textit{e.g.}{#1}}
\long\def\comment#1{}
\usepackage{xspace}
\newcommand{\fixme}[1]{\textcolor{blue}{\xspace\textbf{luo:}\xspace#1}\xspace}
\newcommand{\summer}[1]{\textcolor{red}{\xspace\xspace#1}\xspace}

\begin{document}
\title{Bi-Real Net: Enhancing the Performance of 1-bit CNNs With Improved Representational Capability and Advanced Training Algorithm} 
% Replace with your title

\titlerunning{Bi-Real Net: Enhancing the Performance of 1-bit CNNs}
% Replace with a meaningful short version of your title
%
\author{Zechun Liu\inst{1} \comment{\orcidID{0000-0001-5856-2506}} \and
Baoyuan Wu\inst{2} \comment{\orcidID{0000-0003-2183-5990}} \and
Wenhan Luo\inst{2} \and
Xin Yang\inst{3}\thanks{Corresponding author.} \and
Wei Liu\inst{2} \and
Kwang-Ting Cheng\inst{1}}
%
%Please write out author names in full in the paper, i.e. full given and family names. 
%If any authors have names that can be parsed into FirstName LastName in multiple ways, please include the correct parsing, in a comment to the volume editors:
%\index{Lastnames, Firstnames}
%(Do not uncomment it, because you may introduce extra index items if you do that, we will use scripts for introducing index entries...)
\authorrunning{Zechun Liu et al.}
% Replace with shorter version of the author list. If there are more authors than fits a line, please use A. Author et al.
%

\institute{Hong Kong University of Science and Technology \and
Tencent AI lab \and
Huazhong University of Science and Technology\\
\email{zliubq@connect.ust.hk, \{wubaoyuan1987, whluo.china\}@gmail.com, xinyang2014@hust.edu.cn, wliu@ee.columbia.edu, timcheng@ust.hk}}
\maketitle              % typeset the header of the contribution
\begin{abstract}
In this work, we study the 1-bit convolutional neural networks (CNNs), of which both the weights and activations are binary. While being efficient, the classification accuracy of the current 1-bit CNNs is much worse compared to their counterpart real-valued CNN models on the large-scale dataset, like ImageNet. To minimize the performance gap between the 1-bit and real-valued CNN models, we propose a novel model, dubbed Bi-Real net, which connects the real activations (after the 1-bit convolution and/or BatchNorm layer, before the sign function) to activations of the consecutive block, through an identity shortcut. Consequently, compared to the standard 1-bit CNN, the representational capability of the Bi-Real net is significantly enhanced and the additional cost on computation is negligible. Moreover, we develop a specific training algorithm including three technical novelties for 1-bit CNNs. Firstly, we derive a tight approximation to the derivative of the non-differentiable sign function with respect to activation. Secondly, we propose a magnitude-aware gradient with respect to the weight for updating the weight parameters. Thirdly, we pre-train the real-valued CNN model with a clip function, rather than the ReLU function, to better initialize the Bi-Real net. Experiments on ImageNet show that the Bi-Real net with the proposed training algorithm achieves 56.4\% and 62.2\% top-1 accuracy with 18 layers and 34 layers, respectively. Compared to the state-of-the-arts ({\it e.g.}, XNOR Net), Bi-Real net achieves up to 10\% higher top-1 accuracy with more memory saving and lower computational cost. \footnote{Code is available on: https://github.com/liuzechun/Bi-Real-net}
%\keywords{1-bit CNNs, Bitwise Convolution, Short-cut }
\end{abstract}

\section{Introduction}
Deep Convolutional Neural Networks (CNNs) have achieved substantial advances in a wide range of vision tasks, such as object detection and recognition \cite{alexnet,vggnet,googlenet,resnet,rcnn,faster-rcnn}, depth perception \cite{unsupervised-cnn-depth,monocular-depth}, visual relation detection \cite{zhang2017relation-1,zhang2017relation-2}, face tracking and alignment \cite{facial-detection1,facial-localization2,facial-alignment,wu-face-tracking-iccv,wu-face-tracking-pr}, object tracking \cite{luo2018end}, etc.
However, the superior performance of CNNs usually requires powerful hardware with abundant computing and memory resources. For example, high-end Graphics Processing Units (GPUs). 
Meanwhile, there are growing demands to run vision tasks, such as augmented reality and intelligent navigation, on mobile hand-held devices and small drones. Most mobile devices are not equipped with a powerful GPU neither an adequate amount of memory to run and store the expensive CNN model. 
Consequently, the high demand for computation and memory becomes the bottleneck of deploying the powerful \comment{capability of }CNNs on most mobile devices. 
In general, there are three major approaches to alleviate this limitation. The first is to reduce the number of weights, such as Sparse CNN \cite{sparsecnn}. The second is to quantize the weights (\eg, QNN \cite{qnn} and DoReFa Net \cite{dorefanet}). The third is to quantize both weights and activations, with the extreme case of both weights and activations being binary. 

In this work, we study the extreme case of the third approach, {\it i.e.}, the binary CNNs. 
It is also called 1-bit CNNs, as each weight parameter and activation can be represented by 1-bit. 
As demonstrated in \cite{xnornet},  up to $32 \times$ memory saving and $58 \times$ speedup on CPUs have been achieved for a 1-bit convolution layer, in which the computationally heavy matrix multiplication operations become light-weighted bitwise XNOR operations and bit-count operations. The current binarization method achieves comparable accuracy to real-valued networks on small datasets ({\it e.g.}, CIFAR-10 and MNIST). However on the large-scale datasets ({\it e.g.}, ImageNet), the binarization method based on AlexNet in \cite{binarynet} encounters severe accuracy degradation, {\it i.e.}, from $56.6\%$ to $27.9\%$ \cite{xnornet}. It reveals that the capability of conventional 1-bit CNNs is not sufficient to cover great diversity in large-scale datasets like ImageNet. Another binary network called XNOR-Net \cite{xnornet} was proposed to enhance the performance of 1-bit CNNs, by utilizing the absolute mean of weights and activations.

\begin{figure}[t]
\centering
\includegraphics[width=1\linewidth]{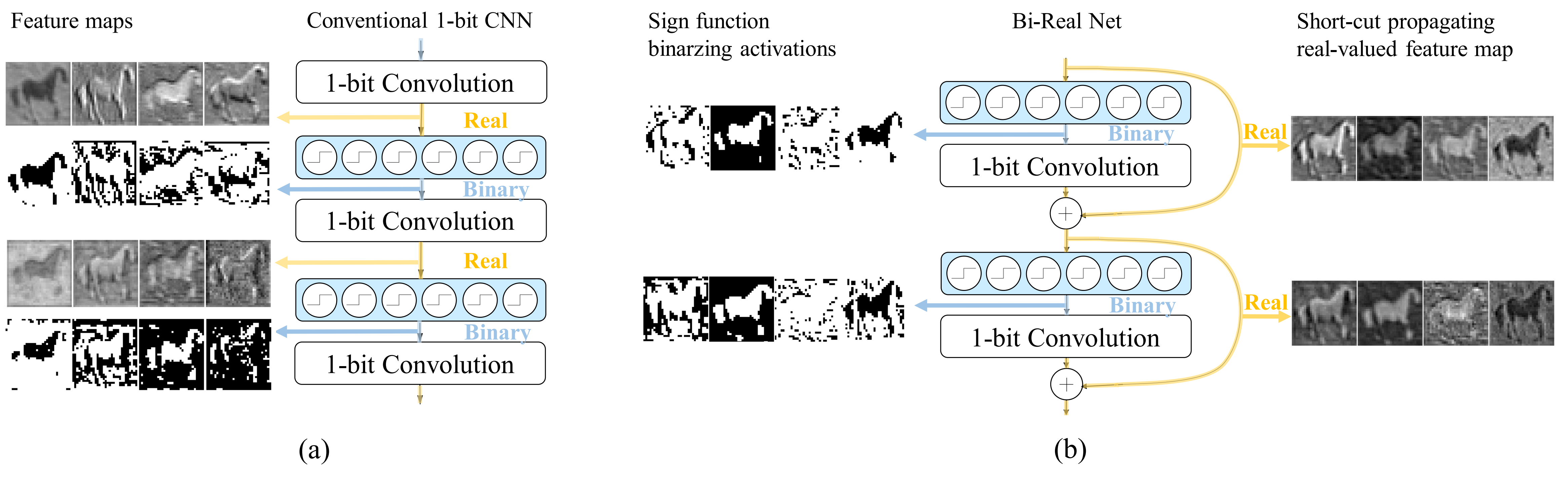}
%\vspace{-0.6cm}
\caption{Network with intermediate feature visualization, yellow lines denote value propagated inside the path being real while blue lines denote binary values. (a) 1-bit CNN without shortcut (b) proposed Bi-Real net with shortcut propagating the real-valued features. }
\label{fig:shortcut_or_not}
%\vspace{-0.4cm}
\end{figure}

The objective of this study is to further improve 1-bit CNNs, as we believe its potential has not been fully explored. 
One important observation is that during the inference process, 1-bit convolution layer generates integer outputs, due to the bit-count operations. The integer outputs will become real values if there is a BatchNorm \cite{batchnorm} layer. But these real-valued activations are then binarized to $-1$ or $+1$ through the consecutive sign function, as shown in Fig. \ref{fig:shortcut_or_not}(a).
Obviously, compared to binary activations, these integers or real activations contain more information, which is lost in the conventional 1-bit CNNs \cite{binarynet}. 
Inspired by this observation, we propose to keep these real activations via adding a simple yet effective shortcut, dubbed Bi-Real net. As shown in Fig. \ref{fig:shortcut_or_not}(b), the shortcut connects the real activations to an addition operator with the real-valued activations of the next block. By doing so, the representational capability of the proposed model is much higher than that of the original 1-bit CNNs, with only a negligible computational cost incurred by                the extra element-wise addition \comment{with a negligible computational cost which only incurs an extra element-wise addition}and without any additional memory cost. 

Moreover, we further propose a novel training algorithm for 1-bit CNNs including three technical novelties:

\begin{itemize}
\item \textbf{Approximation to the derivative of the sign function with respect to activations.} As the sign function binarizing the activation is non-differentiable, we propose to approximate its derivative by a piecewise linear function in the backward pass, derived from the piecewise polynomial function that is a second-order approximation of the sign function. In contrast, the approximated derivative using a step function (\ie, $1_{|x|<1}$) proposed in \cite{binarynet} is derived from the clip function (\ie, clip(-1,x,1)), which is also an approximation to the sign function. We show that the piecewise polynomial function is a closer approximation to the sign function than the clip function. Hence, its derivative is more effective than the derivative of the clip function.

\item \textbf{Magnitude-aware gradient with respect to weights.} As the gradient of loss with respect to the binary weight is not large enough to change the sign of the binary weight, the binary weight cannot be directly updated using the standard gradient descent algorithm. In BinaryNet \cite{binarynet}, the real-valued weight is first updated using gradient descent, and the new binary weight is then obtained through taking the sign of the updated real weight. However, we find that the gradient with respect to the real weight is only related to the sign of the current real weight, while independent of its magnitude. To derive a more effective gradient, we propose to use a magnitude-aware sign function during training, then the gradient with respect to the real weight depends on both the sign and the magnitude of the current real weight. After convergence, the binary weight (\ie, -1 or +1) is obtained through the sign function of the final real weight for inference. 

\item \textbf{Initialization.} As a highly non-convex optimization problem, the training of 1-bit CNNs is likely to be sensitive to initialization. In [17], the 1-bit CNN model is initialized using the real-valued CNN model with the ReLU function pre-trained on ImageNet. We propose to replace ReLU by the clip function in pre-training, as the activation of the clip function is closer to the binary activation than that of ReLU. %The initialization with the clip function \summer{should be} more suitable for 1-bit CNNs than that with ReLU.

\end{itemize}

Experiments on ImageNet show that the above three ideas are useful to train 1-bit CNNs, including both Bi-Real net and other network structures. Specifically, their respective contributions to the\comment{relative} improvements of top-1 accuracy are up to 12\%, 23\% and 13\% for a 18-layer Bi-Real net.
With the dedicatedly-designed shortcut and the proposed optimization techniques, our Bi-Real net, with only binary weights and activations inside each 1-bit convolution layer, achieves 56.4\% and 62.2\% top-1 accuracy with 18-layer and 34-layer structures, respectively, with up to 16.0$\times$ memory saving and 19.0$\times$ computational cost reduction compared to the full-precision CNN. Comparing to the state-of-the-art model (\eg, XNOR-Net), Bi-Real net achieves 10\% higher top-1 accuracy on the 18-layer network.

\section{Related Work}
\textbf{Reducing the number of parameters.} Several methods have been proposed to compress neural networks by reducing the number of parameters and neural connections. For instance, He et al. \cite{resnet} proposed a bottleneck structure which consists of three convolution layers of filter size 1$\times$1, 3$\times$3 and 1$\times$1  with a shortcut connection as a preliminary building block to reduce the number of parameters and to speed up training. In SqueezeNet \cite{squeezenet}, some 3$\times$3 convolutions are replaced with 1$\times$1 convolutions, resulting in a 50$\times$ reduction in the number of parameters. FitNets \cite{fitnets} imitates the soft output of a large teacher network using a thin and deep student network, and in turn yields 10.4$\times$ fewer parameters and similar accuracy to a large teacher network on the CIFAR-10 dataset. In Sparse CNN \cite{sparsecnn}, a sparse matrix multiplication operation is employed to zero out more than 90\% of parameters to accelerate the learning process. Motivated by the Sparse CNN, Han et al. proposed Deep Compression \cite{deepcompression} which employs connection pruning, quantization with retraining and Huffman coding to reduce the number of neural connections, thus, in turn, reduces the memory usage. 

%\vspace{5pt}
\noindent\textbf{Parameter quantization.} The previous study \cite{fwfa} demonstrated that real-valued deep neural networks such as AlexNet \cite{alexnet}, GoogLeNet \cite{googlenet} and VGG-16 \cite{vggnet} only encounter marginal accuracy degradation when quantizing 32-bit parameters to 8-bit. In Incremental Network Quantization, Zhou et al. \cite{incremental} quantize the parameter incrementally and show that it is even possible to further reduce the weight precision to 2-5 bits with slightly higher accuracy than a full-precision network on the ImageNet dataset. In BinaryConnect \cite{binaryconnect}, Courbariaux et al. employ 1-bit precision weights (1 and -1) while maintaining sufficiently high accuracy on the MNIST, CIFAR10 and SVHN datasets.

Quantizing weights properly can achieve considerable memory savings with little accuracy degradation. However, acceleration via weight quantization is limited due to the real-valued activations (\ie, the input to convolution layers). 

Several recent studies have been conducted to explore new network structures and/or training techniques for quantizing both weights and activations while minimizing accuracy degradation. Successful attempts include DoReFa-Net \cite{dorefanet} and QNN \cite{qnn}, which explore neural networks trained with 1-bit weights and 2-bit activations, and the accuracy drops by 6.1\% and 4.9\% respectively on the ImageNet dataset compared to the real-valued AlexNet. Additionally, BinaryNet \cite{binarynet} uses only\comment{ -1 and 1 in all convolution layers } 1-bit weights and 1-bit activations in a neural network and achieves comparable accuracy as full-precision neural networks on the MNIST and CIFAR-10 datasets. In XNOR-Net \cite{xnornet}, Rastegari et al. further improve BinaryNet by multiplying the absolute mean of the weight filter and activation with the 1-bit weight and activation to improve the accuracy. ABC-Net \cite{dji} proposes to enhance the accuracy by using more weight bases and activation bases. The results of these studies are encouraging, but admittedly, due to the loss of precision in weights and activations, the number of filters in the network (thus the algorithm complexity) grows in order to maintain high accuracy, which offsets the memory saving and speedup of binarizing the network.

In this study, we aim to design 1-bit CNNs aided with a real-valued shortcut to compensate for the accuracy loss of binarization. Optimization strategies for overcoming the gradient dismatch problem and discrete optimization difficulties in 1-bit CNNs, along with a customized initialization method, are proposed to fully explore the potential of 1-bit CNNs with its limited resolution.
\begin{figure}[t]
\centering
\includegraphics[width=9cm]{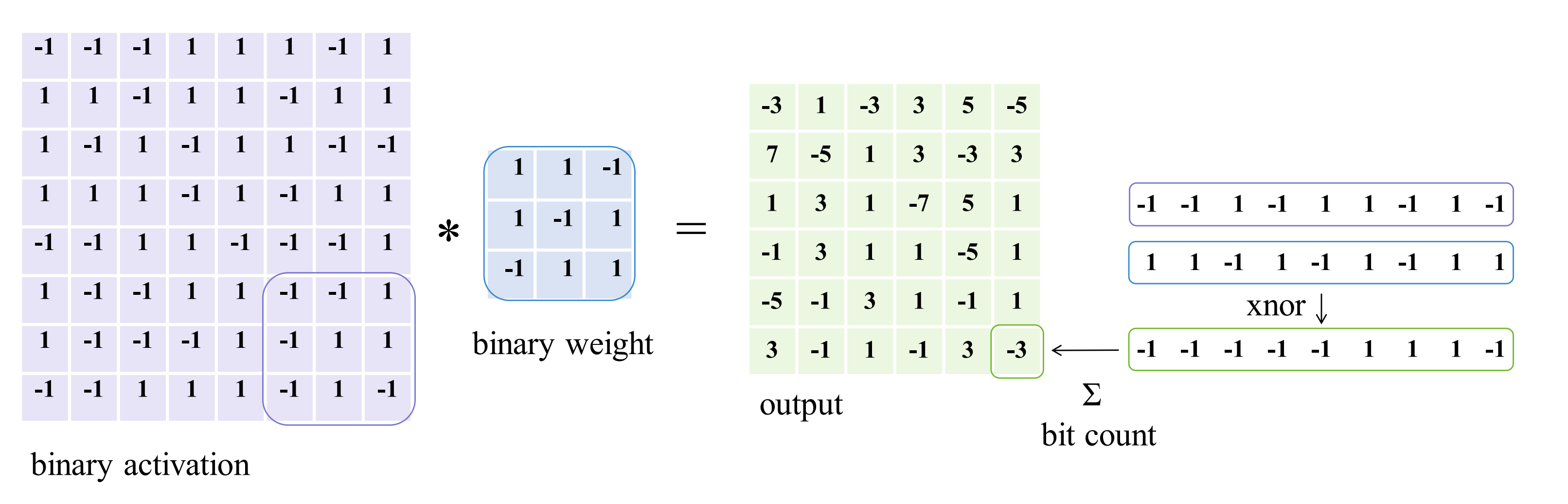}
%\vspace{-0.4cm}
\caption{The mechanism of xnor operation and bit-counting inside the 1-bit CNNs presented in \cite{xnornet}.}
\label{fig:bitcount}
%\vspace{-0.4cm}
\end{figure}

\section{Methodology}
\subsection{Standard 1-bit CNNs and Its Representational Capability}
\begin{figure}[t] 
\centering
\includegraphics[width=1\linewidth]{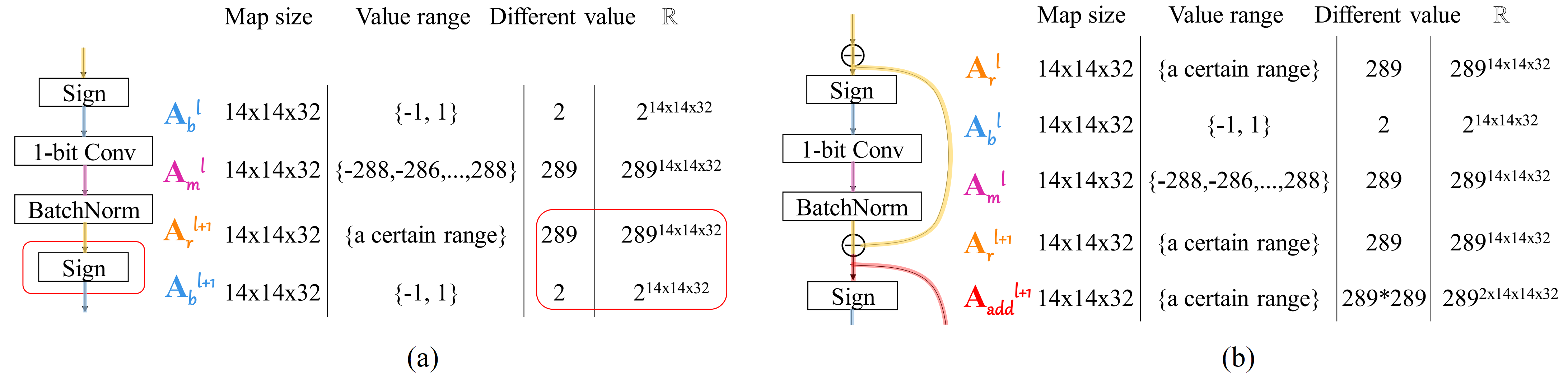}
%\vspace{-0.8cm}
\caption{The representational capability ($\mathbb{R}$) of each layer in (a) 1-bit CNNs without shortcut (b) 1-bit CNNs with shortcut. $\A_b^l$ indicates the output of the Sign function; $\A_m^l$ denotes the output of the 1-bit convolution layer; $\A_r^{l+1}$ represents the output of the BatchNorm layer; The superscript $l$ indicates the block index.}
%\vspace{-0.4cm}
\label{fig:reps}
\end{figure}

1-bit convolutional neural networks (CNNs) refer to the CNN models with binary weight parameters and binary activations in intermediate convolution layers\comment{, while the weight parameters of the first convolution layer and the fully connected layer are still real}. 
Specifically, the binary activation and weight are obtained through a sign function, 
\begin{align}
a_b = {\rm Sign}(a_r) = \left\{  
             \begin{array}{lr}  
             - 1 & {\rm if} \ \ a_r <0 \\  
             + 1 & {\rm otherwise}  
             \end{array}  
\right. 
,
\quad \quad
w_b = {\rm Sign}(w_r) = \left\{  
             \begin{array}{lr}  
             - 1 & {\rm if} \ w_r <0 \\  
             + 1 & {\rm otherwise}  
             \end{array} 
\right. 
,
\label{eq: sign_a and sign_w}
\end{align}
where $a_r$ and $w_r$ indicate the real activation and the real weight, respectively. 
$a_r$ exists in both training and inference process of the 1-bit CNN, due to the convolution and batch normalization (if used). As shown in Fig. \ref{fig:bitcount}, given a binary activation map and a binary $3\times 3$ weight kernel, the output activation could be any odd integer from $-9$ to $9$. If a batch normalization is followed, as shown in Fig. \ref{fig:reps}, then the integer activation will be transformed into real values. 
The real weight will be used to update the binary weights in the training process, which will be introduced later.

Compared to the real-valued CNN model with the 32-bit weight parameter, the 1-bit CNNs obtains up to $32\times$ memory saving. 
Moreover, as the activation is also binary, then the convolution operation could be implemented by the bitwise XNOR operation and a bit-count operation\cite{xnornet}.\comment{\ie, 
\begin{align}
\mathbf{a_b} \cdot \mathbf{w_b}  = \text{bitcount}\text{(XNOR}(\mathbf{a_b},\mathbf{w_b})), % a_i , w_i \in \{ -1, 1 \} \ \forall i 
\label{eq: bitcount}
\end{align}
\comment{where $\mathbf{a_b}$ and $\mathbf{w_b}$ indicate the vectors of binary activations $a_{b,i}$ and binary weights $w_{b,i}$, respectively, with $i$ being the entry index. } }
One simple example of the\comment{ above} bitwise operation is shown in Fig. \ref{fig:bitcount}. 
In contrast, the convolution operation in real-valued CNNs is implemented by the expensive real value multiplication.
Consequently, the 1-bit CNNs could obtain up to 64$\times$ computation saving.

However, it has been demonstrated in \cite{binarynet} that the classification performance of the 1-bit CNNs is much worse than that of the real-valued CNN models on large-scale datasets like ImageNet. 
We believe that the poor performance of 1-bit CNNs is caused by its low representational capacity. 
We denote $\mathbb{R}(\mathbf{x})$ as the representational capability of $\mathbf{x}$, \ie, the number of all possible configurations of $\mathbf{x}$, where $\mathbf{x}$ could be a scalar, vector, matrix or tensor.
For example, the representational capability of 32 channels of a binary $14 \times 14$ feature map $\mathbf{A}$ is $\mathbb{R}(\mathbf{A}) = 2^{14 \times 14 \times 32} = 2^{6272}$. 
Given a $3 \times 3 \times 32$ binary weight kernel $\mathbf{W}$, each entry of $\mathbf{A} \otimes \mathbf{W}$ (\ie, the bitwise convolution output) can choose the even values from (-288 to 288), as shown in Fig \ref{fig:reps}. Thus, $\mathbb{R}(\mathbf{A} \otimes \mathbf{W})$ = $289^{6272}$. Note that since the BatchNorm layer is a unique mapping, it will not increase the number of different choices but scale the (-288,288) to a particular value. If adding the sign function behind the output, each entry in the feature map is binarized, and the representational capability shrinks to $2^{6272}$ again.

\subsection{Bi-Real Net Model and Its Representational Capability}
\comment{As shown in Fig. \ref{fig:reps}, the representational capability of the standard 1-bit CNNs significantly drops from $289^{6272}$ to $2^{6272}$, after the sign function, leading to the significant information loss. }
We propose to preserve the real activations before the sign function to increase the representational capability of the 1-bit CNN, through a simple shortcut.  
Specifically, as shown in Fig. \ref{fig:reps}(b), one block indicates the structure that 
``Sign $\rightarrow$ 1-bit convolution $\rightarrow$ batch normalization $\rightarrow$ addition operator". 
The shortcut connects the input activations to the sign function in the current block to the output activations after the batch normalization in the same block, and these two activations are added through an addition operator, and then the combined activations are inputted to the sign function in the next block. 
 The representational capability of each entry in the added activations is $289^2$. 
 Consequently, the representational capability of each block in the 1-bit CNN with the above shortcut becomes $(289^2)^{6272}$. 
 As both real and binary activations are kept, we call the proposed model as Bi-Real net.

The representational capability of each block in the 1-bit CNN is significantly enhanced due to the simple identity shortcut. 
The only additional cost of computation is the addition operation of two real activations, as these real activations already exist in the standard 1-bit CNN (\ie, without shortcuts). Moreover, as the activations are computed on the fly, no additional memory is needed. 

\begin{figure}[t]
\centering
\includegraphics[width=0.75\linewidth]{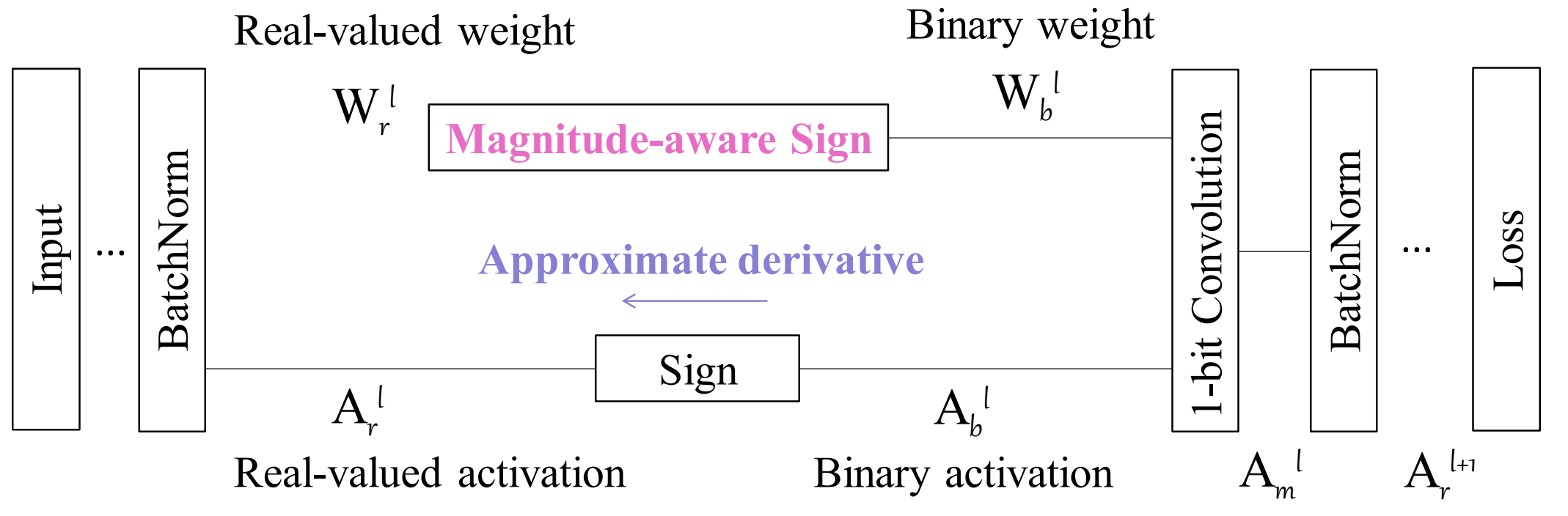}
%\vspace{-0.4cm}
\caption{A graphical illustration of the training process of the 1-bit CNNs, with $A$ being the activation, $W$ being the weight, and the superscript $l$ denoting the $l^{\textit{th}}$ block consisting with Sign, 1-bit Convolution, and BatchNorm. The subscript $r$ denotes real value, $b$ denotes binary value, and $m$ denotes the intermediate output before the BatchNorm layer.}
\label{fig:training}
%\vspace{-0.4cm}
\end{figure}

\subsection{Training Bi-Real Net}
As both activations and weight parameters are binary, the continuous optimization method, \ie, the stochastic gradient descent(SGD), cannot be directly adopted to train the 1-bit CNN. 
There are two major challenges. One is how to compute the gradient of the sign function on activations, which is non-differentiable. The other is that the gradient of the loss with respect to the binary weight is too small to change the weight's sign. 
The authors of \cite{binarynet} proposed to adjust the standard SGD algorithm to approximately train the 1-bit CNN. Specifically, the gradient of the sign function on activations is approximated by the gradient of the piecewise linear function, as shown in Fig. \ref{fig:activation_back}(b). To tackle the second challenge, the method proposed in \cite{binarynet} updates the real-valued weights by the gradient computed with regard to the binary weight and obtains the binary weight by taking the sign of the real weights. 
As the identity shortcut will not add difficulty for training, the training algorithm proposed in \cite{binarynet} can also be adopted to train the Bi-Real net model. 
However, we propose a novel training algorithm to tackle the above two major challenges, which is more suitable for the Bi-Real net model as well as other 1-bit CNNs. 
Besides, we also propose a novel initialization method. 

We present a graphical illustration of the training of Bi-Real net in Fig. \ref{fig:training}. The identity shortcut is omitted in the graph for clarity, as it will not change the main part of the training algorithm. 
\subsubsection{Approximation to the derivative of the sign function with respect to activations.}
\begin{figure}[t]
\centering
\includegraphics[width=12cm]{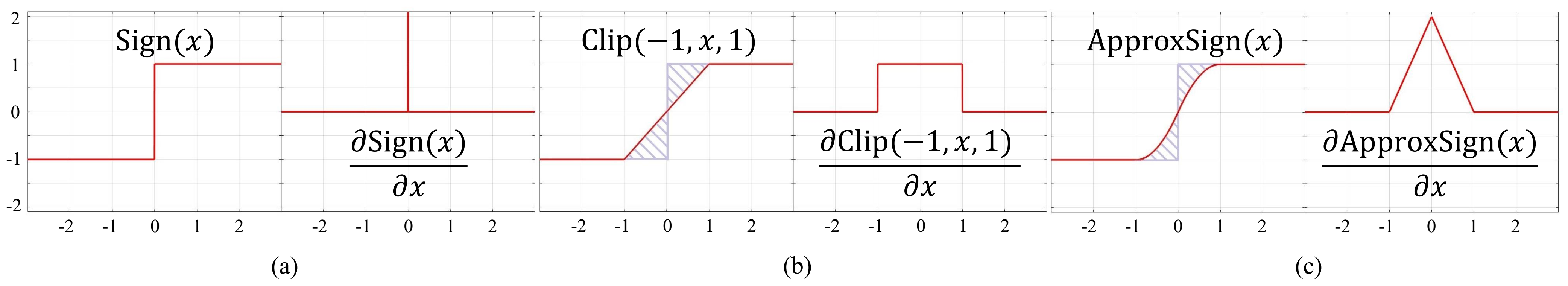}
%\vspace{-0.4cm}
\caption{(a) Sign function and its derivative, (b) Clip function and its derivative for approximating the derivative of the sign function, proposed in \cite{binarynet}, (c) Proposed differentiable piecewise polynomial function and its triangle-shaped derivative for approximating the derivative of the sign function in gradients computation.}
\label{fig:activation_back}
%\vspace{-0.4cm}
\end{figure}

As shown in Fig. \ref{fig:activation_back}(a), the derivative of the sign function is an impulse function, which cannot be utilized in training. 
\comment{\begin{flalign}
\frac{\partial \mathcal{L}}{\partial \mathbf{A}_r^{l,t}} 
= 
\frac{\partial \mathcal{L}}{\partial \mathbf{A}_r^{l+1,t}}  
\frac{\partial \mathbf{A}_r^{l+1,t}}{\partial \mathbf{A}_m^{l,t}} 
\frac{\partial \mathbf{A}_m^{l,t}}{\partial \mathbf{A}_b^{l,t}}\frac{\partial \mathbf{A}_b^{l,t}}{\partial \mathbf{A}_r^{l,t}}
\approx
\frac{\partial \mathcal{L}}{\partial \mathbf{A}_r^{l+1,t}}  
\theta^{l,t}
\mathbf{W}_b^l\frac{\partial F(\mathbf{A}_r^{l,t})}{\partial \mathbf{A}_r^{l,t}}, 
\label{eq: derivative wrt A_r}
\end{flalign}
}
%end comment
\begin{flalign}
\frac{\partial \mathcal{L}}{\partial \mathbf{A}_r^{l,t}} 
= 
\frac{\partial \mathcal{L}}{\partial \mathbf{A}_b^{l,t}}  
\frac{\partial \mathbf{A}_b^{l,t}}{\partial \mathbf{A}_r^{l,t}}
=
\frac{\partial \mathcal{L}}{\partial \mathbf{A}_b^{l,t}}  
\frac{\partial Sign(\mathbf{A}_r^{l,t})}{\partial \mathbf{A}_r^{l,t}}
\approx
\frac{\partial \mathcal{L}}{\partial \mathbf{A}_b^{l,t}}  
\frac{\partial F(\mathbf{A}_r^{l,t})}{\partial \mathbf{A}_r^{l,t}},
\label{eq: derivative wrt A_r}
\end{flalign}
where $F(\mathbf{A}_r^{l,t})$ is a differentiable approximation of the non-differentiable $Sign(\mathbf{A}_r^{l,t})$. 
In \cite{binarynet},  $F(\mathbf{A}_r^{l,t})$ is set as the clip function, leading to the derivative as a step-function  (see \ref{fig:activation_back}(b)).
In this work, we utilize a piecewise polynomial function (see \ref{fig:activation_back}(c)) as the approximation function, as Eq. \eqref{eq4} left.

\begin{align}
\label{eq4}
F(a_r) = \left\{  
             \begin{array}{lr}  
             - 1 & {\rm if} \ a_r  < -1 \\ 
             2a_r+a_r^2 \ \ &{\rm if} -1 \leqslant a_r < 0 \\
             2a_r-a_r^2 &{\rm if} \ 0 \leqslant a_r < 1 \\
             1 & {\rm otherwise}  
             \end{array} 
\right. 
,
\quad 
\frac{\partial F(a_r)}{\partial a_r} = \left\{  
             \begin{array}{lr}  
             2+2a_r \ \ &{\rm if} -1 \leqslant a_r < 0 \\
             2-2a_r &{\rm if} \ 0 \leqslant a_r < 1 \\
             0 & {\rm otherwise}  
             \end{array} 
\right. 
,
\end{align}

\comment{
\begin{align}
\label{eq4}
F(a_r) = \left\{  
             \begin{array}{lr}  
             - 1 & {\rm if} \ a_r  < -1 \\ 
             2a_r+a_r^2 \ \ \ \ &{\rm if} \ -1 \leqslant a_r < 0 \\
             2a_r-a_r^2 &{\rm if} \ 0 \leqslant a_r < 1 \\
             1 & {\rm otherwise}  
             \end{array} 
\right. 
.
\end{align}
}

\noindent
As shown Fig. \ref{fig:activation_back}, the shaded areas with blue slashes can reflect the difference between the sign function and its approximation. The shaded area corresponding to the clip function is $1$, while that corresponding to Eq. \eqref{eq4} left is $\frac{2}{3}$. 
We conclude that Eq. \eqref{eq4} left is a closer approximation to the sign function than the clip function.
Consequently, the derivative of Eq. \eqref{eq4} left is formulated as Eq. \eqref{eq4} right, 
which is a piecewise linear function. 
\comment{
\begin{align}
\label{eq5}
\frac{\partial F(a_r)}{\partial a_r} = \left\{  
             \begin{array}{lr}  
             2+2a_r \ \ \ \ &{\rm if} \ -1 \leqslant a_r < 0 \\
             2-2a_r &{\rm if} \ 0 \leqslant a_r < 1 \\
             0 & {\rm otherwise}  
             \end{array} 
\right. 
,
\end{align}}
%It corresponds to the term above the purple arrow shown in Fig. \ref{fig:training}. 
%We propose to use Eq. \eqref{eq5} to approximate the derivative of the sign function, \ie, $\frac{\partial A_b}{\partial A_r}$. 

\subsubsection{Magnitude-aware gradient with respect to weights.}

Here we present how to update the binary weight parameter in the $l^{th}$ block, \ie, $\mathbf{W}_b^l \in \{-1, +1\}$\comment{, as shown in Fig. \ref{fig:training}}. 
For clarity, we assume that there is only one weight kernel, \ie, $\mathbf{W}_b^l$ is a matrix. 
\comment{Following the gradient descent algorithm, the update is formulated as 
\begin{flalign}
\label{eq6}
\mathbf{W}_b^{l, t+1} = Sign\big( \mathbf{W}_b^{l,t} - \eta \frac{\partial \mathcal{L}}{\partial \mathbf{W}_b^{l,t}} \big),
\end{flalign}
where $t$ denotes the training iteration. 
However, in practice, we find that $\mathbf{W}_b^l$ is rarely changed, as the gradient is not large enough to change the sign in most cases. Obviously, the standard gradient descent Eq. \eqref{eq6} is not suitable for updating the binary weight in the 1-bit CNN.}

The standard gradient descent algorithm cannot be directly applied as the gradient is not large enough to change the binary weights.
To tackle this problem, the method of \cite{binarynet} introduced a real weight 
$\mathbf{W}_r^l$ and a sign function during training. Hence the binary weight parameter can be seen as the output to the sign function, \ie,  $\mathbf{W}_b^l = Sign(\mathbf{W}_r^l)$, as shown in the upper sub-figure in Fig. \ref{fig:training}. 
Consequently, $\mathbf{W}_r^l$ is updated using gradient descent in the backward pass, 
as follows
\begin{flalign}
\mathbf{W}_r^{l, t+1} = 
\mathbf{W}_r^{l,t} - \eta \frac{\partial \mathcal{L}}{\partial \mathbf{W}_r^{l,t} }
=
\mathbf{W}_r^{l,t} - \eta \frac{\partial \mathcal{L}}{\partial \mathbf{W}_b^{l,t}} \frac{\partial \mathbf{W}_b^{l,t}}{\partial \mathbf{W}_r^{l,t} }.
\label{eq: update for W_r}
\end{flalign}
%Then the binary weight is updated using  $\mathbf{W}_b^{l,t+1} = Sign(\mathbf{W}_r^{l,t+1})$. 
Note that $\frac{\partial \mathbf{W}_b^{l,t}}{\partial \mathbf{W}_r^{l,t} }$ indicates the element-wise derivative\comment{, leading to a matrix}. 
In \cite{binarynet}, $\frac{\partial \mathbf{W}_b^{l,t}(i,j)}{\partial \mathbf{W}_r^{l,t}(i,j) }$ is set to $1$ if $\mathbf{W}_r^{l,t}(i,j) \in [-1,1]$, otherwise $0$. 
The derivative $\frac{\partial \mathcal{L}}{\partial \mathbf{W}_b^{l,t}}$ is derived from the chain rule, as follows
\begin{flalign}
\frac{\partial \mathcal{L}}{\partial \mathbf{W}_b^{l,t}} 
= 
\frac{\partial \mathcal{L}}{\partial \mathbf{A}_r^{l+1,t}}  
\frac{\partial \mathbf{A}_r^{l+1,t}}{\partial \mathbf{A}_m^{l,t}} 
\frac{\partial \mathbf{A}_m^{l,t}}{\partial \mathbf{W}_b^{l,t}}
=
\frac{\partial \mathcal{L}}{\partial \mathbf{A}_r^{l+1,t}}  
\theta^{l,t}
\mathbf{A}_b^l, 
\label{eq: derivative wrt W_b}
\end{flalign}
where $\theta^{l,t} = \frac{\partial \mathbf{A}_r^{l+1,t}}{\partial \mathbf{A}_m^{l,t}}$ denotes the derivative of the BatchNorm layer (see Fig. \ref{fig:training}) and has a negative correlation to $\W_b^{l,t}$ . As $\mathbf{W}_b^{l,t} \in \{-1, +1 \}$,  the gradient $\frac{\partial \mathcal{L}}{\partial \mathbf{W}_r^{l,t} }$ is only related to the sign of $\mathbf{W}_r^{l,t}$, while is independent of its magnitude. 

Based on this observation, we propose to replace the above sign function by a magnitude-aware function, as follows:
\begin{flalign}
 \overline{\W}_b^{l,t} =  \frac{ \parallel \W_r^{l,t} \parallel_{1,1}}{|\W_r^{l,t}|} Sign(\W_r^{l,t}),
\label{eq: bar_W_b}
\end{flalign}
where $|\W_r^{l,t}|$ denotes the number of entries in $\W_r^{l,t}$. 
Consequently, the update of $\mathbf{W}_r^l$ becomes 
\begin{flalign}
\mathbf{W}_r^{l, t+1} 
= 
\mathbf{W}_r^{l,t} - \eta \frac{\partial \mathcal{L}}{\partial \overline{\W}_b^{l,t}} \frac{\partial \overline{\W}_b^{l,t}}{\partial \mathbf{W}_r^{l,t} }
=
\mathbf{W}_r^{l,t} - \eta 
\frac{\partial \mathcal{L}}{\partial \mathbf{A}_r^{l+1,t}}  
\overline{\theta}^{l,t}
\mathbf{A}_b^l
\frac{\partial \overline{\W}_b^{l,t}}{\partial \mathbf{W}_r^{l,t} },
\label{eq: new update for W_r}
\end{flalign}
where $\frac{\partial \overline{\W}_b^{l,t}}{\partial \mathbf{W}_r^{l,t} } \approx  \frac{\parallel \W_r^{l,t} \parallel_{1,1}}{|\W_r^{l,t}|} \cdot \frac{\partial Sign(\W_r^{l,t})}{\partial \mathbf{W}_r^{l,t}} \approx \mathbf{1}_{|\W_r^{l,t}|<1}$
\comment{$= \frac{ Sign(\W_r^{l,t}) }{|\W_r^{l,t}|} = \W_r^{l,t} \cdot$} and $\overline{\theta}^{l,t}$ is associated with the magnitude of $\W_r^{l,t}$.
Thus, the gradient $\frac{\partial \mathcal{L}}{\partial \mathbf{W}_r^{l,t} }$ is related to both the sign and magnitude of $\mathbf{W}_r^{l,t}$.
After training for convergence, we still use $Sign(\mathbf{W}_r^l)$ to obtain the binary weight $\W_b^l$ (\ie, -1 or +1), and use $\theta^{l}$ to absorb $\frac{ \parallel \W_r^{l} \parallel_{1,1}}{|\W_r^{l}|}$  and to  associate with the magnitude of $\mathbf{W}_b^{l}$ used for inference.

\subsubsection{Initialization.}
In \cite{dji}, the initial weights of the 1-bit CNNs are derived from the corresponding real-valued CNN model\comment{(\ie, the same model structure as the 1-bit CNNs, except that the sign function on activation is replaced by ReLU)} pre-trained on ImageNet. However, the activation of ReLU is non-negative, while that of Sign is $-1$ or $+1$. Due to this difference,\comment{the weight of} the real CNNs with ReLU may not provide a suitable initial point for training the 1-bit CNNs. Instead, we propose to replace ReLU with $\text{clip}(-1,x,1)$ to pre-train the real-valued CNN model, as the activation of the clip function is closer to the sign function than ReLU. The efficacy of this new initialization will be evaluated in experiments.

\section{Experiments}
In this section, we firstly introduce the dataset for experiments and implementation details in Sec \ref{sec:dataset_implementation}. Then we conduct ablation study in Sec. \ref{sec:ablation_study} to investigate the effectiveness of the proposed techniques. This part is followed by comparing our Bi-Real net with other state-of-the-art binary networks regarding accuracy in Sec \ref{sec:accuracy_comparison}. Sec. \ref{sec:efficiency_comparison} reports memory usage and computation cost in comparison with other networks.

\subsection{Dataset and Implementation Details}
\label{sec:dataset_implementation}
The experiments are carried out on the ILSVRC12 ImageNet classification dataset \cite{imagenet}. ImageNet is a large-scale dataset with 1000 classes and 1.2 million training images and 50k validation images. Compared to other datasets like CIFAR-10 \cite{cifar10} or MNIST \cite{mnist}, ImageNet is more challenging due to its large scale and great diversity. The study on this dataset will validate the superiority of the proposed Bi-Real network structure and the effectiveness of three training methods for 1-bit CNNs. In our comparison, we report both the top-1 and top-5 accuracies.

For each image in the ImageNet dataset, the smaller dimension of the image is rescaled to 256 while keeping the aspect ratio intact. For \textit{training}, a random crop of size 224 $\times$ 224 is selected. Note that, in contrast to XNOR-Net and the full-precision ResNet, we do not use the operation of random resize, which might improve the performance further. For \textit{inference}, we employ the 224 $\times$ 224 center crop from images. 

\comment{\noindent\summer{\textbf{Pre-training:} We prepare the real-valued network for initializing binary network in three steps: 1) Train the network with ReLU nonlinearity function from scratch, following the hyper-parameter settings in \cite{resnet}. 2) Replace ReLU with SReLU\cite{srelu} with the range of (-1,1) and the negative slope of 0.1 and finetune the network for 20 epochs. 3) Finetune the network with clip(-1,x,1) nonlinearity instead of SReLU for 12 epochs. \fixme{this does not deserve such a long paragraph. Better to descript it in the training paragraph.}
}}

\noindent\textbf{Training:} We train two instances of the Bi-Real net, including an \textit{18-layer Bi-Real net} and a \textit{34-layer Bi-Real net}.  The training of them consists of two steps: training the 1-bit convolution layer and retraining the BatchNorm. In the first step, the weights in the 1-bit convolution layer are binarized to the sign of real-valued weights multiplying the absolute mean of each kernel. We use the SGD solver with the momentum of 0.9 and set the weight-decay to 0, which means we no longer encourage the weights to be close to 0. For the 18-layer Bi-Real net, we run the training algorithm for 20 epochs with a batch size of 128. The learning rate starts from 0.01 and is decayed twice by multiplying 0.1 at the 10\textit{th} and the 15\textit{th} epoch. For the 34-layer Bi-Real net, the training process includes 40 epochs and the batch size is set to 1024. The learning rate starts from 0.08 and is multiplied by 0.1 at the 20\textit{th} and the 30\textit{th} epoch, respectively. In the second step, we constraint the weights to -1 and 1, and set the learning rate in all convolution layers to 0 and retrain the BatchNorm layer for 1 epoch to absorb the scaling factor.

\noindent\textbf{Inference:} we use the trained model with binary weights and binary activations in the 1-bit convolution layers for inference.

\begin{figure}[t]
\label{fig:three_building_blocks}
\centering
\includegraphics[height=2.5cm]{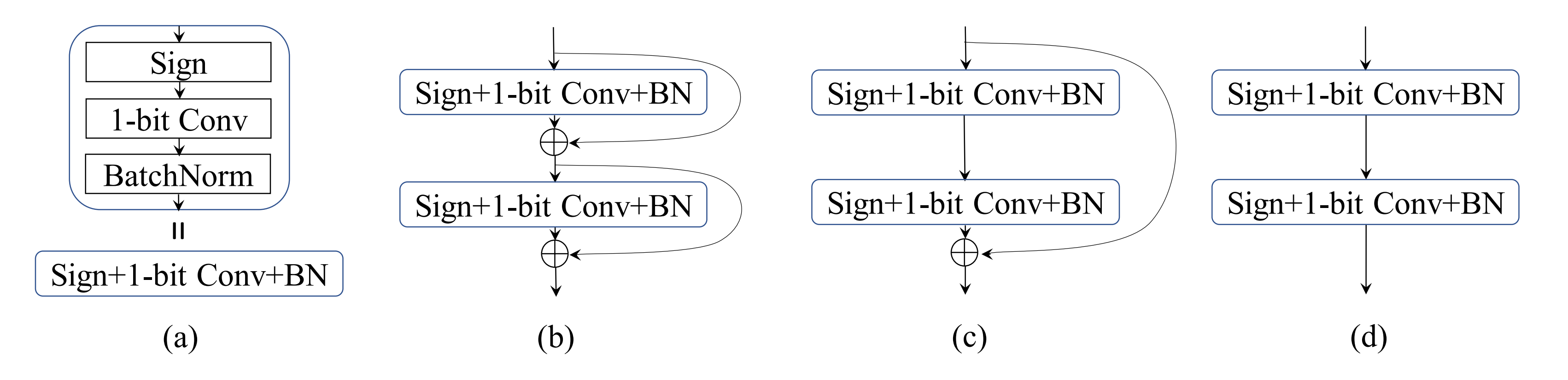}
%\vspace{-0.4cm}
\caption{Three different networks differ in the shortcut design of connecting the blocks shown in (a) conjoint layers of Sign, 1-bit Convolution, and the BatchNorm. (b) Bi-Real net with shortcut bypassing every block (c) Res-Net with shortcut bypassing two blocks, which corresponds to the ReLU-only pre-activation proposed in \cite{identity_mapping} and (d) Plain-Net without the shortcut. These three structures shown in (b), (c) and (d) have the same number of weights.}
%\vspace{-0.4cm}
\label{fig:compare_structure}
\end{figure}

\subsection{Ablation Study}
\label{sec:ablation_study}
\noindent\textbf{Three building blocks.} The shortcut in our Bi-Real net transfers real-valued representation without additional memory cost, which plays an important role in improving its capability. To verify its importance, we implemented a Plain-Net structure without shortcut as shown in Fig. \ref{fig:compare_structure} (d) for comparison. At the same time, as our network structure employs the same number of weight filters and layers as the standard ResNet, we also make a comparison with the standard ResNet shown in Fig. \ref{fig:compare_structure} (c). For a fair comparison, we adopt the ReLU-only pre-activation ResNet structure in \cite{identity_mapping}, which differs from Bi-Real net only in the structure of two layers per block instead of one layer per block. The layer order and shortcut design in Fig. \ref{fig:compare_structure} (c) are also applicable for 1-bit CNN. The comparison can justify the benefit of implementing our Bi-Real net by specifically replacing the 2-conv-layer-per-block Res-Net structure with two 1-conv-layer-per-block Bi-Real structure.

As discussed in Sec. 3, we proposed to overcome the optimization challenges induced by discrete weights and activations by 1) approximation to the derivative of the sign function with respect to activations, 2) magnitude-aware gradient with respect to weights and 3) clip initialization. To study how these proposals benefit the 1-bit CNNs individually and collectively, we train the 18-layer structure and the 34-layer structure with a combination of these techniques on the ImageNet dataset. Thus we derive $2 \times 3 \times 2 \time 2 \times 2 \times 2= 48$ pairs of values of top-1 and top-5 accuracy, which are presented in Table \ref{table:ablation study}.

\setlength{\tabcolsep}{1pt}
\begin{table}[t]
\scriptsize
\begin{center}
\caption{Top-1 and top-5 accuracies (in percentage) of different combinations of the three proposed techniques on three different network structures, Bi-Real net, ResNet and Plain Net, shown in Fig.\ref{fig:compare_structure}.}
\label{table:ablation study}
%\vspace{-0.2cm}
\begin{tabular}{lllllllllllllll}
\hline
\noalign{\smallskip}
Initiali- & Weight &Activation & \multicolumn{2}{c}{Bi-Real-18} &  \multicolumn{2}{c}{Res-18} & \multicolumn{2}{c}{Plain-18} &\multicolumn{2}{c}{Bi-Real-34} &\multicolumn{2}{c}{Res-34} &\multicolumn{2}{c}{Plain-34}  \\
zation&update&backward&top-1&top-5&top-1&top-5&top-1&top-5&top-1&top-5&top-1&top-5&top-1&top-5\\
\noalign{\smallskip}
\hline
\noalign{\smallskip}
\multirow{5}{*}{ReLU} & \multirow{2}{*}{Original} &Original&\cellcolor{Gray}32.9& \cellcolor{Gray}56.7& 27.8& 50.5& 3.3& 9.5& \cellcolor{Gray}53.1& \cellcolor{Gray}76.9& 27.5& 49.9& 1.4& 4.8\\
\cline{3-15}
\noalign{\smallskip}
& &Proposed&\cellcolor{Gray}36.8& \cellcolor{Gray}60.8& 32.2& 56.0& 4.7& 13.7& \cellcolor{Gray}58.0& \cellcolor{Gray}81.0& 33.9& 57.9& 1.6& 5.3\\
\cline{2-15}
\noalign{\smallskip}
&\multirow{2}{*}{Proposed} &Original&\cellcolor{Gray}40.5& \cellcolor{Gray}65.1& 33.9& 58.1& 4.3& 12.2& \cellcolor{Gray}59.9& \cellcolor{Gray}82.0& 33.6& 57.9& 1.8& 6.1\\
\cline{3-15}
\noalign{\smallskip}
&&Proposed&\cellcolor{Gray}47.5& \cellcolor{Gray}71.9& 41.6& 66.4& 8.5& 21.5& \cellcolor{Gray}61.4& \cellcolor{Gray}83.3& 47.5& 72.0& 2.1& 6.8\\
\cline{2-15}
\noalign{\smallskip}
&\multicolumn{2}{l}{Real-valued Net} &\cellcolor{Gray}68.5& \cellcolor{Gray}88.3& 67.8& 87.8& 67.5& 87.5& \cellcolor{Gray}70.4& \cellcolor{Gray}89.3& 69.1& 88.3& 66.8& 86.8\\
\hline
\noalign{\smallskip}
\multirow{5}{*}{Clip} & \multirow{2}{*}{Original} &Original&\cellcolor{Gray}37.4& \cellcolor{Gray}62.4& 32.8& 56.7& 3.2& 9.4& \cellcolor{Gray}55.9& \cellcolor{Gray}79.1& 35.0& 59.2& 2.2& 6.9\\
\cline{3-15}
\noalign{\smallskip}
& &Proposed&\cellcolor{Gray}38.1& \cellcolor{Gray}62.7& 34.3& 58.4& 4.9& 14.3& \cellcolor{Gray}58.1& \cellcolor{Gray}81.0& 38.2& 62.6& 2.3& 7.5\\
\cline{2-15}
\noalign{\smallskip}
&\multirow{2}{*}{Proposed} &Original&\cellcolor{Gray}53.6& \cellcolor{Gray}77.5& 42.4& 67.3& 6.7& 17.1& \cellcolor{Gray}60.8& \cellcolor{Gray}82.9& 43.9& 68.7& 2.5& 7.9\\
\cline{3-15}
\noalign{\smallskip}
&&Proposed&\cellcolor{Gray}\textbf{56.4}& \cellcolor{Gray}\textbf{79.5}& 45.7& 70.3& 12.1& 27.7& \cellcolor{Gray}\textbf{62.2}& \cellcolor{Gray}\textbf{83.9}& 49.0& 73.6& 2.6& 8.3\\
\cline{2-15}
\noalign{\smallskip}
&\multicolumn{2}{l}{Real-valued Net} &\cellcolor{Gray}68.0& \cellcolor{Gray}88.1& 67.5& 87.6& 64.2& 85.3& \cellcolor{Gray}69.7& \cellcolor{Gray}89.1& 67.9& 87.8& 57.1& 79.9\\
\hline
\noalign{\smallskip}
\multicolumn{5}{l}{Full-precision original ResNet\cite{resnet}}& 69.3& 89.2&&&&& 73.3 &91.3\\
\hline
\end{tabular}
%\vspace{-0.8cm}
\end{center}
\end{table}
\setlength{\tabcolsep}{1.4pt}

Based on Table \ref{table:ablation study}, we can evaluate each technique's individual contribution and collective contribution of each unique combination of these techniques towards the final accuracy.

1) Comparing the $4^{\textit{th}}-7^{\textit{th}}$ columns with the $8^{\textit{th}}-9^{\textit{th}}$  columns, both the proposed Bi-Real net and the binarized standard ResNet outperform their plain counterparts with a significant margin, which validates the effectiveness of shortcut and the disadvantage of directly concatenating the 1-bit convolution layers. As Plain-18 has a thin and deep structure, which has the same weight filters but no shortcut, binarizing it results in very limited network representational capacity in the last convolution layer, and thus can hardly achieve good accuracy. 

2) Comparing the $4^{\textit{th}}-5^{\textit{th}}$  and $6^{\textit{th}}-7^{\textit{th}}$ columns, the 18-layer Bi-Real net structure improves the accuracy of the binarized standard ResNet-18 by about 18\%. This validates the conjecture that the Bi-Real net structure with more shortcuts further enhances the network capacity compared to the standard ResNet structure. Replacing the 2-conv-layer-per-block structure employed in Res-Net with two 1-conv-layer-per-block structure, adopted by Bi-Real net, could even benefit a real-valued network.

3) All proposed techniques for initialization, weight update and activation backward improve the accuracy at various degrees. For the 18-layer Bi-Real net structure, the improvement from the weight (about 23\%, by comparing the $2^{\textit{nd}}$ and $4^{\textit{th}}$ rows) is greater than the improvement from the activation (about 12\%, by comparing the $2^{\textit{nd}}$ and $3^{\textit{rd}}$ rows) and the improvement from replacing ReLU with Clip for initialization (about 13\%, by comparing the $2^{\textit{nd}}$ and $7^{\textit{th}}$ rows).  These three proposed training mechanisms are independent and can function collaboratively towards enhancing the final accuracy.

4) The proposed training methods can improve the final accuracy for all three networks in comparison with the original training method, which implies these proposed three training methods are universally suitable for various networks. 

5) The two implemented Bi-Real nets (\textit{i.e.} the 18-layer and 34-layer structures) together with the proposed training methods, achieve approximately 83\% and 89\% of the accuracy level of their corresponding full-precision networks, but with a huge amount of speedup and computation cost saving. 

\textit{In short}, the shortcut enhances the network representational capability, and the proposed training methods help the network to approach the accuracy upper bound.
%\vspace{-0.2cm}
\setlength{\tabcolsep}{1pt}
\begin{table}[t]
\begin{center}
\caption{This table compares both the top-1 and top-5 accuracies of our Bi-real net with other state-of-the-art binarization methods: BinaryNet \cite{binarynet} , XNOR-Net \cite{xnornet}, ABC-Net \cite{dji} on both the Res-18 and Res-34 \cite{resnet}. The Bi-Real net outperforms other methods by a considerable margin.}
%\vspace{-0.3cm}
\label{table:accuracy_comparison}
\begin{tabular}{cccccccc}
\hline\noalign{\smallskip}
& & Bi-Real net & BinaryNet & ABC-Net  & XNOR-Net  & Full-precision \\
\noalign{\smallskip}
\hline
\multirow{2}{*}{18-layer}  & \ Top-1 &  56.4\%  & 42.2\% & 42.7\% & 51.2\% & 69.3\% \\
 & \ Top-5 & 79.5\%   & 67.1\% & 67.6\% & 73.2\% & 89.2\% \\
\hline
\multirow{2}{*}{34-layer} & \ Top-1 & 62.2\%   & -- & 52.4\% & -- &73.3\% \\
& \ Top-5 & 83.9\%  & -- & 76.5\% & -- &91.3\% \\
\hline
\end{tabular}
%\vspace{-0.8cm}
\end{center}
\end{table}
\setlength{\tabcolsep}{1.4pt}

\subsection{Accuracy Comparison With State-of-The-Art}
%\vspace{-0.1cm}
\label{sec:accuracy_comparison}
While the ablation study demonstrates the effectiveness of our 1-layer-per-block structure and the proposed techniques for optimal training, it is also necessary to compare with other state-of-the-art methods to evaluate Bi-Real net's overall performance. To this end, we carry out a comparative study with three  methods: BinaryNet \cite{binarynet}, XNOR-Net \cite{xnornet} and ABC-Net \cite{dji}. These three networks are representative methods of binarizing both weights and activations for CNNs and achieve the state-of-the-art results. Note that, for a fair comparison, our Bi-Real net contains the same amount of weight filters as the corresponding ResNet that these methods attempt to binarize, differing only in the shortcut design.

Table \ref{table:accuracy_comparison} shows the results. The results of the three networks are quoted directly from the corresponding references, except that the result of BinaryNet is quoted from ABC-Net \cite{dji}. 
The comparison clearly indicates that the proposed Bi-Real net outperforms the three networks by a considerable margin in terms of both the top-1 and top-5 accuracies. Specifically, the 18-layer Bi-Real net outperforms its 18-layer counterparts BinaryNet and ABC-Net with relative 33\% advantage, and achieves a roughly 10\% relative improvement over the XNOR-Net. Similar improvements can be observed for 34-layer Bi-Real net. In short, our Bi-Real net is more competitive than the state-of-the-art binary networks.

\subsection{Efficiency and Memory Usage Analysis}
\label{sec:efficiency_comparison}
In this section, we analyze the saving of memory usage and speedup in computation of Bi-Real net by comparing with the XNOR-Net \cite{xnornet} and the full-precision network individually.

The memory usage is computed as the summation of 32 bit times the number of real-valued parameters and 1 bit times the number of binary parameters in the network. For efficiency comparison, we use FLOPs to measure the total real-valued multiplication computation in the Bi-Real net, following the calculation method in \cite{resnet}. As the bitwise XNOR operation and bit-counting can be performed in a parallel of 64 by the current generation of CPUs, the FLOPs is calculated as the amount of real-valued floating point multiplication plus 1/64 of the amount of 1-bit multiplication.

\setlength{\tabcolsep}{1pt}
\begin{table}[t]
\begin{center}
\caption{Memory usage and FLOPs calculation in Bi-Real net.}
%\vspace{-0.2cm}
\label{table:memoy_n_flop}
\begin{tabular}{ccccccc}
\hline\noalign{\smallskip}
& & Memory usage \ & \ Memory saving &  \ FLOPs &  \ Speedup \\
\noalign{\smallskip}
\hline
\noalign{\smallskip}
\multirow{3}{*}{18-layer}  & \ Bi-Real net &  33.6 Mbit & 11.14 $\times$ &1.63 $\times 10^8$ & 11.06 $\times$  \\
 & \ XNOR-Net & 33.7 Mbit & 11.10 $\times$ &1.67 $\times 10^8$ & 10.86 $\times$ \\
 & \ Full-precision Res-Net & 374.1 Mbit & -- &1.81 $\times 10^9$ & --\\
\hline
\multirow{3}{*}{34-layer}  & \ Bi-Real net & 43.7 Mbit & 15.97 $\times$ &1.93 $\times 10^8$ & 18.99 $\times$  \\
 & \ XNOR-Net & 43.9 Mbit & 15.88 $\times$ &1.98 $\times 10^8$ & 18.47 $\times$  \\
 & \ Full-precision Res-Net & 697.3 Mbit & -- &3.66 $\times 10^9$ & --\\

\hline
\end{tabular}
\vspace{-0.5cm}
\end{center}
\end{table}
\setlength{\tabcolsep}{1.4pt}

We follow the suggestion in XNOR-Net \cite{xnornet}, to keep the weights and activations in the first convolution and the last fully-connected layers to be real-valued. We also adopt the same real-valued 1x1 convolution in Type B short-cut \cite{resnet} as implemented in XNOR-Net. Note that this 1x1 convolution is for the transition between two stages of ResNet and thus all information should be preserved. As the number of weights in those three kinds of layers accounts for only a very small proportion of the total number of weights, the limited memory saving for binarizing them does not justify the performance degradation caused by the information loss.

For both the 18-layer and the 34-layer networks, the proposed Bi-Real net reduces the memory usage by 11.1 times and 16.0 times individually, and achieves computation reduction of about 11.1 times and 19.0 times, in comparison with the full-precision network. \comment{This memory saving and speedup is comparable with the XNOR-Net, while our network achieves higher precision than it.} 
Without using real-valued weights and activations for scaling binary ones during inference time, our Bi-Real net requires fewer FLOPs and uses less memory than XNOR-Net and is also much easier to implement.

%\summer{Without using real-valued weights and activations for scaling binary ones during inference time, our Bi-Real net requires fewer FLOPs and uses less memory than XNOR-Net and is also much easier to implement.}

\section{Conclusion}
In this work, we have proposed a novel 1-bit CNN model, dubbed Bi-Real net. Compared with the standard 1-bit CNNs, Bi-Real net utilizes a simple short-cut to significantly enhance the representational capability. Further, an advanced training algorithm is specifically designed for training 1-bit CNNs (including Bi-Real net), including a tighter approximation of the derivative of the sign function with respect the activation, the magnitude-aware gradient with respect to the weight, as well as a novel initialization. 
Extensive experimental results demonstrate that the proposed Bi-Real net and the novel training algorithm show superiority over the state-of-the-art methods. 
In future, we will explore other advanced integer programming algorithms (\eg, Lp-Box ADMM \cite{wu2018lp}) to train Bi-Real Net. 

%We propose a Bi-Real net structure with three novel training techniques to optimize the 1-bit CNNs for which both activations and weights in the intermediate convolution layers are binarized. By propagating the real-valued intermediate representation with short-cuts throughout the Bi-Real net, the learning capability can be greatly boosted. 

%method of approximation to the derivative of the sign function with respect to activations, the magnitude-aware gradient with respect to weights and the initialization with clip non-linearity outperform the original training method and can work collaboratively towards improving the final accuracy.

\bibliographystyle{splncs04}
\bibliography{egbib}
%\newpage
\comment{

}
\end{document}